# Towards Accurate Word Segmentation for Chinese Patents

Si Li[1], Nianwen Xue[2]

[1]Beijing University of Posts and Telecommunications, Beijing, China.
[2]Brandeis University, Massachusetts, USA.
Email: lisi@bupt.edu.cn, xuen@brandeis.edu

A patent is a property right for an invention granted by the government to the inventor. An invention is a solution to a specific technological problem. So patents often have a high concentration of scientific and technical terms that are rare in everyday language. The Chinese word segmentation model trained on currently available everyday language data sets performs poorly because it cannot effectively recognize these scientific and technical terms. In this paper we describe a pragmatic approach to Chinese word segmentation on patents where we train a character-based semi-supervised sequence labeling model by extracting features from a manually segmented corpus of 142 patents, enhanced with information extracted from the Chinese TreeBank. Experiments show that the accuracy of our model reached 95.08% ($F_1$ score) on a held-out test set and 96.59% on development set, compared with an $F_1$ score of 91.48% on development set if the model is trained on the Chinese TreeBank. We also experimented with some existing domain adaptation techniques, the results show that the amount of target domain data and the selected features impact the performance of the domain adaptation techniques.

## Introduction

Patents are exclusive rights granted by a sovereign state to an inventor in exchange for detailed public disclosure of an invention. By analyzing large amounts of patent data, one can potentially gain insights into new technological trends for purposes of technology forecasting or competitor monitoring. With the large number of patent filings, it is increasingly hard for human analysts to manually examine the patents to identify technological trends, and there is a pressing need to develop Natural Language Processing techniques to automate the process. This article is concerned with the issue of processing Chinese patents with natural language techniques, which has its unique challenges. It is well known that Chinese text does not come with natural word delimiters, and the first step for many Chinese language processing tasks is word segmentation, the automatic determination of word boundaries in Chinese text. Tremendous progress was made in this area in the last decade or so due to the availability of large-scale human segmented corpora coupled with better statistical modeling techniques. On the data side, there exist a few large-scale human annotated corpora based on established word segmentation standards, and these include the Chinese TreeBank (Xue et al. 2005), the Sinica Balanced Corpus (Chen et al. 1996), the PKU Peoples' Daily Corpus (Duan et al. 2003), and the LIVAC balanced corpus (T'sou et al. 1997) developed in mainland China, Hong Kong, Taiwan and the United States. These corpora were used in a series of international Chinese word segmentation bake-offs (http://www.sighan.org/) that played a crucial role in advancing the state of the art in Chinese word segmentation. Another driver for the improvement in Chinese word segmentation accuracy comes from the evolution of statistical modeling techniques. Dictionaries used to play a central role in early heuristics-based word segmentation techniques such as the maximum match, where entries in a dictionary are used to match strings in an unsegmented input sentence (Chen and Liu 1992). Their role was affirmed in statistical finite-state models (Sproat et al. 1996) where dictionaries are used to build segmentation graphs for a sentence and statistics are then used to search for the best word segmentation path. Modern word segmentation systems have moved away from dictionary-based approaches in favor of character tagging approaches, where each character is assigned a label indicating its position within a word. This allows the word segmentation problem to be modeled as a sequence labeling problem, and lends itself to advanced discriminative sequence modeling techniques such as Maximum Entropy Markov models (Xue 2003) and Conditional Random Fields (Peng et al. 2004) that can take advantage of a large feature space. More recently, perceptron learning based systems also produced very competitive performance (Zhang and Clark 2007). With these better modeling techniques, state-of-the-art systems routinely report accuracy in the high 90 percentage points, with a few recent systems reporting accuracies of over 98% in $F_1$ score (Sun 2011; Zeng et al. 2013b).

Chinese word segmentation is far from being a solved problem however and significant challenges remain. Advanced word segmentation systems perform very well in domains such as newswire where there is a large amount of human annotated training data. There is often a rapid degradation in performance when systems trained on one domain (let us call it the *source* domain) are used to segment data in a different domain (let us call it the *target* domain), especially when the target domain is distant from the source domain. Technical documentation such as patents is one such domain where there is relatively little human annotated data that can be used to train supervised statistical machine learning systems. In our effort to develop an accurate word segmentation system for Chinese patents, we took the following three pragmatic steps. (1) We first manually annotated a corpus of 142 patents which contain about 440K words following the Chinese TreeBank word segmentation standards (Xia 2000) to train and test our word segmentation models. (2) We then developed a number of new features that are more adaptable to new domains or more portable across domains. In particular, we propose a group of novel document-level features based on the writing style of patents and show that these new features further improve the word segmentation accuracy for patents. We also use a set of character similarity features that prove to be very portable across domains. (3) Finally, we experimented with a few existing domain adaptation techniques in an attempt to further improve the accuracy of our Chinese word segmentation system. Domain adaptation is essentially a way of making use of out-of-domain data to improve the performance of a system in a target domain. In this case, the out-of-domain data set we use is the Chinese TreeBank data, which we try to use to improve the word segmentation accuracy of patent data. Evaluated on the patent data set we annotated, our system is able to achieve an accuracy of 96.59% ($F_1$ score) on the development set and 95.08% ($F_1$ score) on the test set.

In addition to successfully developing an accurate Chinese word segmentation system for patents, we also made several significant findings. The first is that even a smaller in-domain data training set is significantly better than a much larger out-of-domain training set. We also found that domain adaptation techniques interact in complex ways with the actual features used in the word segmentation model as well as the size of the in-domain training set. Domain adaptation techniques are most useful when the in-domain training set is sufficiently small. This suggests that there are limitations to existing domain adaptation techniques and we cannot just use a domain adaptation technique blindly without considering the amount of target domain training data and the features of the model.

The rest of this article is organized as follows. Section Method describes our proposed patent word segmentation model in detail. In Section Experiments, we discuss the characteristics of the patent data and present our experimental results. We review the related work in Section Related. Finally, Section Conclusion concludes the article.

# Method

We adopt the character-based sequence labeling approach, first proposed in (Xue 2003), as our modeling technique for its simplicity and effectiveness. This approach treats each sentence as a sequence of characters and assigns to each character a label that indicates its position within a word. In this paper, we use the *BMES* tag set to indicate the character positions. The tag set has four labels that represent four possible positions a character can occupy within a word: *B* for beginning, *M* for middle, *E* for ending, and *S* for a single character as a word. After each character in a sentence is tagged with a *BMES* label, a sequence of words can be derived from this labeled character sequence.

We train a Conditional Random Field (CRF) (Lafferty et al. 2001) model for this sequence labeling problem. When extracting features to train a CRF model from a sequence of *n* characters $C_1C_2...C_{i-1}C_iC_{i+1}...C_n$, we extract features for each character $C_i$ from a fixed window of characters. We start with a set of core features extracted from the annotated corpus that have been shown to be effective in previous work and propose some new features tailored to patent word segmentation. The features are grouped into three categories: character-based features as baseline features, in-domain document-level features which are obtained from patent data by considering each patent document independently, and out-of-domain features which are obtained by using the information from source domain data. We describe each group of features in detail below.

*Character features* (*CF*)

When predicting the position of a character within a word, features based on its surrounding characters and their types have shown to be the most effective features for this task (Xue 2003). There are some variations of these features depending on the window size in terms of the number of characters to examine, and here we adopt the feature templates used in (Ng and Low 2004).

*Character N-gram features.* The N-gram features are various combinations of the characters surrounding the candidate character $C_i$. The 10 features we used are listed below:
- Character unigrams: $C_k$ (*i*-3<*k*<*i*+3)
- Character bigrams: $C_kC_{k+1}$ (*i*-3<*k*<*i*+2) and $C_{k-1}C_{k+1}$ (*k*=*i*)

*Character type N-gram features.* We classify the characters in Chinese text into 4 types: numbers (Arabic numerals and Chinese numerals), Chinese characters or *hanzi* (exclude Chinese numerals), English letters, and others. $T_i$ is the character type of $C_i$. The character type has been used in the previous work in various forms (Ng and Low 2004; Jiang et al. 2009), and the 4 features we use are as follows:
- Character type unigrams: $T_k$ (*k*=*i*)
- Character type bigrams: $T_kT_{k+1}$ (*i*-2<*k*<*i*+1) and $T_{k-1}T_{k+1}$ (*k*=*i*)

Starting with this baseline, we extract some new features to improve Chinese patent word segmentation accuracy.

*In-domain document-level features*

A patent is a property right for an invention granted by the government to the inventor, and many of the patents have a high concentration of scientific and technical terms. From a machine learning perspective, these terms are hard to detect and segment because they are often "new words" that are not seen in everyday language. These technical terminologies also tend to be very sparse, either because they are related to the latest invention that has not made into everyday language, or because our limited patent dataset cannot possibly cover all possible technical topics. However, these technical terms are also topical and they tend to have high relative frequency within a patent document even though they are sparse in the entire patent data set. We attempt to exploit this distribution property with some document-level features which are extracted from each patent document.

*Longest n-gram features* (*LNG*). We propose a longest n-gram (LNG) feature as a document-level feature. Each patent document is treated as an independent unit and the candidate longest n-gram sequence lists for each patent are obtained as described in Algorithm 1.

---

**ALGORITHM 1:** Longest n-gram sequence extraction.

**Input:**
Sentences $\{s_i\}$ in patent $P_i$;

**Output:**
Longest n-gram sequence list for $P_i$;

1: **For** each sentence $s_i$ in $P_i$ **do**:
n-gram sequence extraction ($2 \leq n \leq length(s_i)$);
2: Count the frequency of each n-gram sequence;
3: Delete the sequence if its frequency<2;
4: Delete sequence $i$ if it is contained in a longer sequence $j$;
5: All the remaining sequences form a longest n-gram sequence list for $P_i$;
6: **return** Longest n-gram sequences list.

---

For a given patent, the LNG feature value for the target character $C_i$'s LNG is set to 'S' if the bigram ($C_i$, $C_{i+1}$) are the first two characters of an n-gram sequence in this patent's longest n-gram sequence list. If ($C_{i-1}$, $C_i$) are the last two characters of an n-gram sequence in this patent's longest n-gram sequence list, the target character $C_i$'s LNG is set to 'F'. It is set to 'O' otherwise. If $C_i$ can be labeled as both 'S' and 'F' at the same time, label 'T' will be given as the final label. For example, if 'α' is the target character $C_i$ in patent A and the sequence 'α－干扰素' is in patent A's longest n-gram sequence list. If the character next to 'α' is '－', the value of the LNG feature is set to 'S'. If the next character is not '－', the value of the LNG feature is set to 'O'.

*Pseudo Kullback-Leibler divergence* (*PKL*). The second document-level feature we propose is the pseudo Kullback-Leibler divergence feature which is calculated following the form of the Kullback-Leibler divergence. The Kullback-Leibler divergence measures the difference between two probability distributions $P$ and $Q$ by considering the same variable. For pseudo Kullback-Leibler divergence, we use the marginal probability distribution functions of two different variables.

The relative position information is very important for Chinese word segmentation as a sequence labeling task. Characters *XY* may constitute a meaningful word, but characters *YX* may not be. Therefore, if we want to determine whether character *X* and character *Y* can form a word, the relative position of these two characters should be considered. We adopt a pseudo KL divergence with the relative position information as a measure of the association strength between two adjacent characters *X* and *Y*. The pseudo KL divergence is an asymmetric measure. The *PKL* value between character *X* and character *Y* is described in Algorithm 2.

---

**ALGORITHM 2:** Pseudo KL divergence.

**Input:**
    Sentences $\{s_i\}$ in patent $P_i$;

**Output:**
    Pseudo KL divergence values between different characters in $P_i$;

1:   **For** each sentence $s_i$ in $P_i$ **do**:
       trigram sequences extraction;
2:   Count the frequency of each trigram;
3:   Delete the trigram if its frequency<2;
4:   **For** $C_i$ in trigram $C_iC_{i+1}C_{i+2}$ **do** :

$$PKL(C_i, C_{i+1}) = p(C_i^1) \log \frac{p(C_i^1)}{p(C_{i+1}^2)} \tag{1}$$

$$PKL(C_i, C_{i+2}) = p(C_i^1) \log \frac{p(C_i^1)}{p(C_{i+2}^3)} \tag{2}$$

    The superscripts {1,2,3} indicate the character position in trigram sequences;
5:   **return** $PKL(C_i, C_{i+1})$ and $PKL(C_i, C_{i+2})$ for the first character $C_i$ in each trigram.

---

The *PKL* values are real numbers and are sparse. A common solution to sparsity reduction is binning. We rank the *PKL* values between two adjacent characters in each patent from low to high, and then divide all values into five bins. Each bin is assigned a unique ID and all *PKL* values in the same bin are replaced by this ID. This ID is then used as the *PKL* feature value for the target character $C_i$.

*Pointwise Mutual information* (*PMI*). Pointwise Mutual information has been widely used in previous work on Chinese word segmentation (Sun and Xu 2011; Zhang et al. 2013b) and it is a measure of the mutual dependence of two strings and reflects the tendency of two strings appearing in one word. In previous work, PMI statistics are gathered on the entire data set, and here we gather PMI statistics for each patent in an attempt to capture character strings with high PMI in a particular patent. The procedure for calculating PMI is the same as that for computing pseudo KL divergence, but the functions (1) and (2) are replaced with the following functions:

$$PMI(C_i, C_{i+1}) = \log \frac{p(C_i^1, C_{i+1}^2)}{p(C_i^1)p(C_{i+1}^2)} \tag{3}$$

$$PMI(C_i, C_{i+2}) = \log \frac{p(C_i^1, C_{i+2}^3)}{p(C_i^1)p(C_{i+2}^3)} \tag{4}$$

For the target character $C_i$, we obtain the values for *PMI*($C_i$, $C_{i+1}$) and *PMI*($C_i$, $C_{i+2}$). In each patent document, we rank these values from high to low and divide them into five bins. Then the PMI feature values are represented by the bin IDs.

*Out-of-domain features*

When we train a word segmentation model, training data that is not from the domain of the test data is considered to be out-of-domain and is not expected to be as useful as in-domain training data, that is, data that is in the same domain as the test data. Still, out-of-domain data may share some common characteristics with the in-domain training set that can be exploited. Since our patent data is segmented following the Chinese TreeBank segmentation standards, and the Chinese TreeBank is fairly large-corpus consisting of 1.2 million words (Version 7.0), we try to use it as a data source from which certain features can be extracted. The features we extract from the CTB either represent a property of a character (e.g., the POS tag of a character, or if this character is ever part of a multi-character word in the CTB) or relations between characters (the similarity between two characters). These features are not bound to a particular context and can be stored in a dictionary indexed by the characters. They can simply be retrieved from a dictionary when used as features for a target domain.

*Character POS feature* (*C_POS*). Chinese words are composed of Chinese hanzi, and an overwhelming majority of these Chinese characters can be single-character words themselves in some context. In fact, most of the multi-character words are compounds that are 2-4 characters in length. The formation of these compound words is not random and abides by word formation rules that are similar to the formation of phrases (Xue 2000; Packard 2000). For example, the compound noun "地板/floor" is a noun formed by two single-character nouns "地/ground" and "板/board", the compound verb "敬献/present respectfully" is composed of the adverb "敬/respectfully" and the verb "献/present". In fact, the Chinese TreeBank word segmentation guidelines (Xia 2000) specify how words are segmented based on the part-of-speech (POS) of their component characters. We hypothesize that the POS tags of the single-character words would be useful information to help predict how they form compound words, and these POS tags are more fine-grained information than the character type information described in the previous section, but are more robust and generalizable than the characters themselves.

Since we do not have POS-tagged patent data, we extract this information from the Chinese TreeBank. We extract the POS tags for all the single-character words in the CTB. Some of the single-character words will have more than one POS tag. In this case, we select the most frequent POS tag as the C_POS tag for this character. The result of this extraction process is a list of single-character Chinese words, each of which is assigned a single POS tag. When extracting features for the target character $C_i$, if $C_i$ is in this list, the POS tag of $C_i$ is used as a feature for this target character.

*Word Dictionary feature* (*Dict*). Whether or not a character is part of a word in a large dictionary says something about the distributional characteristics of this character. Have a dictionary like this may help us correctly segment words that are in an existing dictionary. We automatically compile this dictionary from the CTB, and when compiling this dictionary, we only select 2-character or 3-character words.
For a given target character $C_i$, if one of the following character sequences appears in the dictionary, the Dict feature is set to 1. Otherwise, it is set to 0. The character combinations are $C_iC_{i+1}C_{i+2}$, $C_{i-1}C_iC_{i+1}$, $C_{i-2}C_{i-1}C_i$, $C_iC_{i+1}$ and $C_{i-1}C_i$.

*Character similarity feature* (*Sim*). The character similarity feature captures the intuition that the similarity characteristics between adjacent characters may tell us something about how they should be segmented. In order to compute the similarity between adjacent characters, we need to first have a vectorial representation of each character based on its distribution in a large corpus. This corpus does not have to be word segmented, but it needs to be sentence segmented because the computation of this distribution crucially relies on two characters occurring in the same sentence. We still use the Chinese TreeBank to compute the distribution of the characters, but obviously we can use any sentence segmented corpus for this purpose. The algorithm for computing the distribution of the characters is presented in algorithm 3.

---

**ALGORITHM 3:** Character distribution matrix.

**Input:**
  Sentences $\{s_i\}$ in the unlabeled data;

**Output:**
  Feature vector matrix $F$;

1: Character unigram set $S$ $\{C_0, C_1, ... C_{i-1}, C_i, C_{i+1}, …C_{n-1}\}$ is set up based on the unlabeled data, the size of $S$ is $n$;
2: Matrix $M_{n\times n}$=[0];
3: **IF** $C_i$ and $C_j$ appear in sentence $s_i$ **do**:
    $M[i][j]=M[i][j]+1$   $0 \leq i,j < n$ and $i \neq j$;
4: Matrix $P_{n\times n}=$ PPMI($M_{n\times n}$);
5: Matrix $F_{n\times k}=$ SVD($P_{n\times n}$);
6: **return** Matrix $F$, in which each row corresponds to a $k$ dimensional feature vector for each character in $S$ as its character distribution.

---

As discussed in (Bollegala et al. 2014), Algorithm 3 computes a feature vector for a character $C_i$ by using unigrams that co-occur with $C_i$ in a sentence. We start by initializing a feature co-occurrence matrix $M$ of dimension of $n \times n$, where $n$ is the number of unique characters in a corpus. The value of each element $e_{ij}$ in $M$ is the number of sentences that is incremented each time when $C_i$ and $C_j$ co-occur in a sentence. Based on this raw sentence frequency count, the Positive Pointwise Mutual Information (PPMI), for each element in $M$ can be computed. PPMI is a variation of PMI. If PMI value is less than zero, the value is set to zero in PPMI (Lin 1998; Bullinaria and Levy 2007). After we compute the PPMI matrix $P$ from $M$, we then apply Singular Value Decomposition (SVD) to matrix $P$ to reduce the dimensionality and obtain a matrix $F$. Each row $i$ in $F$ represents the distribution of character $C_i$ in a $k$ dimensional feature vector. We can use $F$ to compute the similarity of any two characters represented in $F$.

For a given character $C_i$ in the target data set, we compute the similarity between $C_i$ and $C_{i+1}$, $C_{i+2}$, $C_{i-1}$, $C_{i-2}$ respectively, by consulting $F$. So for each target character $C_i$ we will have four similarity values that we use as features. The similarity measure we use is cosine similarity, and if a character $C$ does not have a representation in $F$, we the similarity between this character and any other character to 0.

## Experiments

In this section we present our experimental results. We first describe the data sets we used for our experiments and then present experimental results that the features we proposed are effective for the patent data sets. Finally we present domain adaptation experiments that show the effectiveness of the domain adaptation techniques are tied the size of the target training set as well as the features used, suggesting that there are limitations to these domain adaptation techniques and that they cannot be blindly adopted.

*Data sets*

*Out-of-domain data set*. For out-of-domain data sets we use the Chinese TreeBank (CTB) 7.0. The Chinese TreeBank is a word segmented, POS-tagged and syntactically bracketed corpus and it is widely used in the NLP community to train word segmentation, POS-tagging, and syntactic parsing systems. This version of the Chinese TreeBank consists of 2,448 text files, 51,447 sentences, 1,196,329 words and 1,931,381 hanzi (Chinese characters). This data set has a variety of different sources, including Xinhua news wire, news magazine articles, transcribed broadcast news and broadcast conversations, as well as newsgroup and weblog articles. However, none of these data sources are technical in nature. We use the word segmentation and POS tags annotation in this data set and make no use of its syntactic structures in our experiments.

*In-domain data set*. Since we are not aware of a publicly available manually annotated Chinese patent data sets that we can use for training and benchmarking purposes, we annotated 142 Chinese patents following the CTB word segmentation guidelines (Xia 2000). Since the original guidelines are mainly designed to cover non-technical everyday language, particularly newswire, many scientific and technical terms found in patents are not covered in the guidelines. We had to extend the CTB word segmentation guidelines to handle these new cases. Deciding on how to segment these scientific and technical terms is a big challenge since these patents cover many different technical fields and without proper technical background, even a native speaker has difficulty in segmenting them properly. For example, "大肠杆菌" is a biomedical terminology, and it means "colibacillus", but since "大肠" (meaning "colon") and "杆菌" (meaning "bacillus") are also words in Chinese, there are two possible ways of segmenting the string "大肠杆菌": as two words or as one single word. This is a familiar dilemma in word segmentation of Chinese text, even for everyday language. The difference is that in this case, one has to have some background knowledge in bio-medicine in order to realize that "大肠杆菌" is a technical term and should be treated as one word. For difficult scientific and technical terms, we consult BaiduBaike ("Baidu Encyclopedia", http://baike.baidu.com/), which we use as a scientific and technical terminology dictionary during our annotation. There are still many words that do not appear in BaiduBaiKe, and these include chemical names and formulas. These chemical names and formulas (e.g., "１－溴－３－氯丙烷/1-bromo-3-chloropropane") are usually very long, and unlike everyday words, they often have numbers and punctuation marks in them. We decided not to try segmenting the internal structures of such chemical terms and treat them as single words, because without a technical background in chemistry, it is very hard to segment their internal structures consistently.

The annotated patent dataset covers many topics and they include chemistry, mechanics, medicine, etc. If we consider the words in our *annotated dataset* but not in CTB 7.0 data as *new words* (or out-of-vocabulary, OOV), the new words account for 18.3% of the patent corpus by token and 68.1% by type. This shows that there is a large number of words in the patent corpus that are not in the everyday language vocabulary. Table 1 presents the data split used in our experiments.

Table 1. Training, development and test data on Patent data

| Data set | # of words | # of patent |
|---|---|---|
| Training | 345,336 | 113 |
| Development | 46,196 | 14 |
| Test | 48,351 | 15 |

*Experiments on effectiveness of features*

We use CRF++ (Kudo 2013) to train our sequence labeling model. *Precision*, *recall*, $F_1$ score and $R_{OOV}$ are used to evaluate our word segmentation methods, where $R_{OOV}$ for our purposes means the recall of new words which do not appear in CTB 7.0 but in our newly annotated patent data set. This model is trained and tested on our newly annotated patent data sets, but we use data from CTB 7.0 to extract information as an external knowledge source that we can use to define our features.

Table 2 shows the segmentation results on the patent development set with different feature combinations. The model with the CF feature templates is considered to be the baseline system. We add one feature template at a time to investigate the effectiveness of each feature type. From Table 2, we can see that adding the new features we proposed leads to consistent improvement across all experimental conditions, and that the LNG features are the most effective and bring about the largest improvement in accuracy. Adding the C_POS features to the CF+LNG+PMI+PKL feature combination only leads to a slight improvement in the $R_{OOV}$ rate, but not the $F_1$ score. When all the features that make a positive contribution are added to the model, the final $F_1$ score improves 1.07% absolute and the $R_{OOV}$ rate improves 1.86% absolute over the baseline. $k$ is set at 2000 for the character similarity feature. A similar improvement over the baseline is observed on the test set. The final feature combination (CF+LNG+PMI+PKL+C_POS+Dict+Sim) leads to an improvement of 0.94% absolute over the baseline in $F_1$ score. The improvement in the recall of OOV words is even higher, amounting to 3.11% absolute over the baseline. This shows that improvement from these features are very stable.

Table 2. Segmentation performance by using patent data as training data.

| Train set | Test set | Features | P | R | $F_1$ | $R_{OOV}$ |
|---|---|---|---|---|---|---|
| Patent train | Patent dev. | CF | 95.53 | 95.51 | 95.52 | 90.63 |
| | | CF+C_POS | 95.68 | 95.53 | 95.60 | 90.59 |
| | | CF+LNG | 96.27 | 96.19 | 96.23 | 91.41 |
| | | CF+LNG+PKL | 96.38 | 96.21 | 96.30 | 91.70 |
| | | CF+LNG+PMI | 96.45 | 96.32 | 96.39 | 92.07 |
| | | CF+LNG+PMI+PKL | 96.48 | 96.38 | 96.43 | 92.24 |
| | | CF+LNG+PMI+PKL+C_POS | 96.53 | 96.32 | 96.43 | 92.35 |
| | | CF+LNG+PMI+PKL+C_POS+Dict | 96.53 | 96.34 | 96.44 | 92.36 |
| | | CF+LNG+PMI+PKL+C_POS+Dict+Sim | 96.66 | 96.52 | 96.59 | 92.49 |
| Patent train | Patent test | CF | 93.84 | 94.44 | 94.14 | 85.10 |
| | | CF+LNG+PMI+PKL+C_POS+Dict+Sim | 94.96 | 95.19 | 95.08 | 88.21 |

*Experiments on domain adaptation methods*

A typical scenario for using domain adaptation techniques is when there is limited in-domain annotated data and a large amount of out-of-domain annotated data. Ideally these two data sets are annotated following the same standards. This seems to apply to our situation, where we have a smaller manually segmented patent data set and a much larger out-of-domain data set in the Chinese TreeBank. So we experimented with several commonly used domain adaptation techniques to see if we can further improve the word segmentation accuracy for Chinese patents. It is important to note that domain adaptation techniques are different from using the out-of-domain data sets as a knowledge source for extracting features, the way we have used the Chinese TreeBank to extract out-of-domain features. Domain adaptation techniques typically involves combining the out-of-domain data set (designated as the *source domain data*) and the in-domain data (designated as the *target domain data*) in some way, and use the combined data as training data. For example, a simple and yet effective domain adaptation method, first proposed in (Daumé III 2007), first augments the feature space of both the source and target data and then use the combined feature space to train the target domain model. In our specific case, the source data is the Chinese TreeBank and the target data is the patent data we annotated. This is the main domain adaptation technique that we experiment with, and the reader is referred to (Daumé III 2007) for implementation details. We refer to this method as the 'Easy' method, following (Daumé III 2007). The feature augmentation in the 'Easy' method can be described with Equation (5):

$$\Phi^s(x) = <x, x, 0>, \Phi^t(x) = <x, 0, x> \qquad (5)$$

where *s* and *t* represent the source domain and target domain respectively. Suppose we are deciding on how to label the character "方/square". In the source data it appears in "主办方/sponsor" and in the target data it appears in "方钻杆/Kelly". Assume further that "方钻杆/kelly" never appears in the source data. In the source domain, "方/square" is tagged as `E' and the value of its feature vector <LNG, Dict> is <F, 1>. In the target domain, "方/square" is tagged as 'B' and the value of this feature vector is <S, 0>. After feature augmentation using the 'Easy' method, the feature vectors are changed to <F, 1, F, 1, 0, 0> for source domain data and <S, 0, 0, 0, S, 0> for target data.

We first establish a baseline by training a model on only the CTB data and evaluate the model on the patent development and test set (see Table 1 for the data split). Table 3 shows the evaluation results for this baseline model results. The results show that using the combined features (CF+LNG+PMI+PKL+C_POS+Dict+Sim) improves over the baseline features (CF). The results also show, however, that the model trained on the source data alone performs much worse than the model trained on the target data alone. That shows the importance of having a training set in the same domain.

Table 3. Segmentation performance when using different training data.

| Train set | Test set | Features | $P$ | $R$ | $F_1$ | $R_{OOV}$ |
|---|---|---|---|---|---|---|
| CTB train | Patent dev. | CF | 87.85 | 88.85 | 88.35 | 70.15 |
| | | CF+LNG+PMI+PKL+C_POS+Dict+Sim | 91.38 | 91.59 | 91.48 | 78.10 |
| CTB train | Patent test | CF | 86.10 | 86.30 | 86.20 | 67.55 |
| | | CF+LNG+PMI+PKL+C_POS+Dict+Sim | 89.17 | 88.59 | 88.88 | 72.86 |
| Patent train | Patent dev. | CF | 95.53 | 95.51 | 95.52 | 90.63 |
| | | CF+LNG+PMI+PKL+C_POS+Dict+Sim | 96.66 | 96.52 | 96.59 | 92.49 |
| Patent train | Patent test | CF | 93.84 | 94.44 | 94.14 | 85.10 |
| | | CF+LNG+PMI+PKL+C_POS+Dict+Sim | 94.96 | 95.19 | 95.08 | 88.21 |

We next experimented with domain adaptation techniques to see if there are ways to effectively combine the source data and the target data to further improve the word segmentation accuracy for the target domain. The main domain adaptation method is the 'Easy' method, and we also compare it with three obvious baseline approaches. One is 'Target' only method which only uses the annotated patent data as training data. The second method is the 'All' method which simply concatenate the source data with the target data into one large training set and extract features from it. The third is the 'Transit' method (Jiang et al. 2009; Daumé III and Marcu 2006; Daumé III 2007) which uses the predictions made by the model trained on the source data as features for a model trained on the target data. Specifically, the model trained on the source data is first used to label the target data. Then these labels are used as a feature to the model trained the target training set.

The four methods are listed below:
- Method 1: 'Target' method with Combined features
- Method 2: 'All' method with Combined features
- Method 3: 'Transit' method with Combined features
- Method 4: 'Easy' domain adaptation method with Combined features

To observe the impact of different target training data sizes on the performance of domain adaptation techniques, we use 10 different target training data sizes to train 4 different methods with 'Combined' features (CF+LNG+PMI+PKL+C_POS+Dict+Sim). The 10 data sizes are shown in Table 4.

Table 4. Patent Document Index

| Index | # of words | Index | # of words |
|---|---|---|---|
| 1 | 17,527 | 6 | 184,644 |
| 2 | 35,951 | 7 | 217,426 |
| 3 | 73,158 | 8 | 253,650 |
| 4 | 112,140 | 9 | 290,036 |
| 5 | 149,900 | 10 | 329,067 |

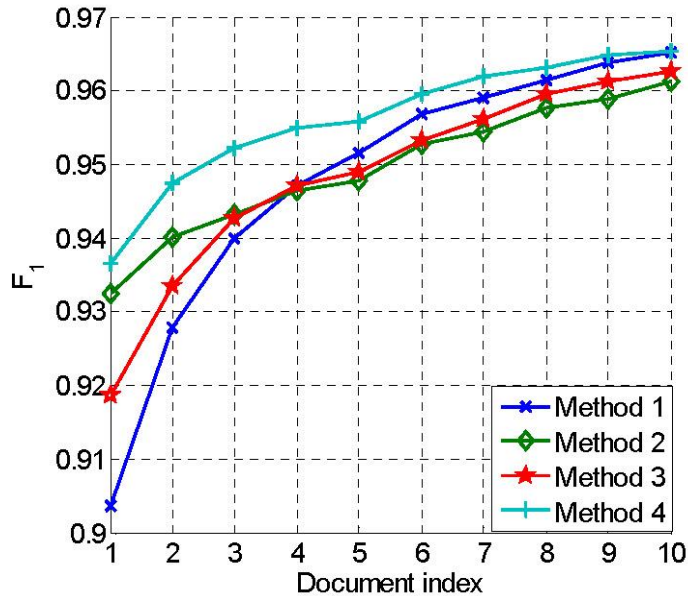

Fig. 1. The $F_1$ score of different methods with different training data size.

Figure 1 plots the $F_1$ scores of the different methods for different target training data sizes. There are a few observations we can make here. The first observation is that all curves in Figure 1 show a rising tendency, indicating that the accuracy of the model improves as the target training set gets bigger. We can also see that when the target training set is small, Methods 2, 3 and 4, which make use of the source training set in addition to the target training set, performs better on $F_1$ score than Method 1, which uses the target training set alone. The third observation is that the 'Easy' domain adaptation technique (Method 4) outperforms the simple concatenation method (Method 2) and the transit method (Method 3) across all target training data sizes. It is the only method that is not overtaken by using the target training set alone as the target training set increases. When the target training set is over 329k words, the 'Easy' domain adaptation method performs virtually the same as the target training set only method. This means that the 'Easy' domain adaptation technique is only effective when the target training data set is small.

Figure 2 plots the OOV *recall* rate and shows a slightly different trend. The 'Easy' domain adaptation technique is quickly overtaken by the target training set only model on the OOV *recall* rate at about 112k words, and after that the target only model is the most effective in correctly segmenting OOV words. Given that the 'Easy' domain adaptation technique performs better or equally well $F_1$ score than the target only method at all target training data sizes, we can conclude that the better OOV *recall* rate is at the expense of lower accuracy for segmenting in-vocabulary words.

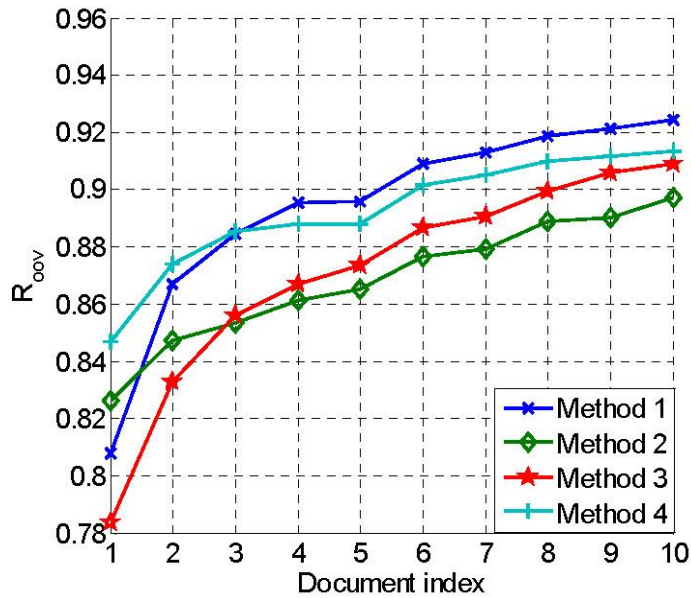

Fig. 2. The $R_{OOV}$ score of different methods with different training data size.

We conducted an additional experiment to investigate the impact of different feature sets on the 'Easy' domain adaptation technique. We repeated the above experiment using a subset of the features (CF+LNG+PMI+PKL+C_POS). This simplify things, we only compare the 'Target' only method and 'Easy' method. They are marked as Method 5 and Method 6 respectively. From Figure 3, we can see that the performance of the Target only method (Method 5) outperforms the 'Easy' method (Method 6) when target training set is over 290k words. This reflects a different tendency than what is observed in Figure 1 when the full feature set is used, and suggests that when we uses domain adaptation techniques we have to pay attention not only to how much target training data we have but also to the actual features in the model.

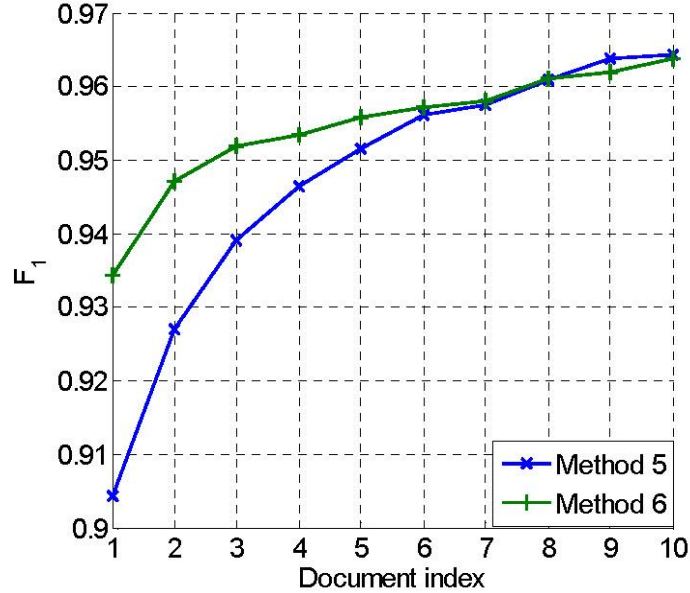

Fig. 3. The $F_1$ score of different methods with CF+LNG+PMI+PKL+C_POS features.

**Related work**

Most of the previous work on Chinese word segmentation focused on news wire, and one widely adopted technique is character-based representation combined with sequential learning models such as Maximum Entropy Markov models (Xue 2003; Low et al. 2005) and Conditional Random Field (Zhao et al. 2006; Sun and Xu 2011; Zeng et al. 2013b; Zhang et al. 2013b; Wang and Kan 2013). More recently, word-based models using perceptron learning techniques (Zhang and Clark 2007) also produce very competitive results. There are also some recent successful attempts to combine character-based and word-based techniques (Sun 2010; Zeng et al. 2013a). The deep learning method is also involved in Chinese word segmentation task (Zheng et al. 2013; Pei et al. 2014).

As Chinese word segmentation has reached a very high accuracy in the newswire domain, the attention of the field has started to shift to other domains where there are few annotated resources and the problem is more challenging. For example, there has been some recent work on the word segmentation of literature data (Liu and Zhang 2012; Liu et al. 2014; Zhang et al. 2014). Work on informal language genres such as microblogs (Wang and Kan 2013; Zhang et al. 2013a) is also starting to emerge. While challenging, these data sources still fall within the scope of "everyday" language that native speakers do not have to have any special training to fully understand.

Patents are distinctly different from the above genres as they contain scientific and technical terms that require some special training to understand. From the point of view of automatic word segmentation, these scientific and technical terms are often "new" words that are difficult to segment because they do not appear in currently available annotated everyday language resources. There has been very little work in this area, and the only work that is devoted to Chinese word segmentation is (Guo et al. 2012), which reports work on Chinese patent word segmentation with a fairly small test set without any annotated training data in the target domain. They reported an accuracy of 86.42% ($F_1$ score), but the results are incomparable with ours as their evaluation data is not available to us. We differ from their work in that we manually segmented a significant amount of data, and trained a model with document-level features designed to capture the characteristics of patent data.

# Conclusion

In this paper, we presented an accurate character-based word segmentation model for Chinese patents. We show that word segmentation models trained on currently available data sets do not work well on patents. We then adopted a pragmatic approach where we first manually annotated a significant amount of patent data, and then designed features to capture the distributional characteristics of the scientific and technical terms in patents. We are able to achieve an accuracy of over 95.08% $F_1$ score on the test set and 96.59% on the development set, compared with 88.88% $F_1$ score on the test set and 91.48% on the development set if the model is trained on the Chinese TreeBank.

Our contributions are three fold. Our first contribution is that we have annotated a significant amount of Chinese patent data that we plan to make publicly available, and this will help other researchers to further improve the word segmentation accuracy on Chinese patents. Our second contribution is that we have proposed novel in-domain and out-of-domain features that prove to be effective in improving the word segmentation accuracy of both patents and an Internet novel data set. Our third contribution is that we have shown that existing domain adaptation techniques interact in complex ways with the size of the target training data and the features used in a model, and this means close attention needs to be paid to the size of the target training set and the selection of features when deciding whether to include data from the source domain.

# References

Danushka Bollegala, David Weir, and John Carroll. 2014. Learning to Predict Distributions of Words Across Domains. In Proceedings of ACL'14. 613–623.

John A. Bullinaria and Joseph P. Levy. 2007. Extracting semantic representations from word co-occurrence statistics: A computational study. Behavior Research Methods (2007), 510–526.

Keh-Jiann Chen, Chu-Ren Huang, Li-Ping Chang, and Hui-Li Hsu. 1996. Sinica Corpus: Design Methodology for Balanced Corpora. In Proceedings of the 11th Pacific Asia Conference on Language, Information and Computation. 167–176.

Keh-Jiann Chen and Shing-Huan Liu. 1996. Word Identification for Mandarin Chinese Sentences. In Proceedings of COLING'92. 101–107.

Hal Daumé III. 2007. Frustratingly easy domain adaptation. In Proceedings of ACL'07. 256–263.

Hal Daumé III and Daniel Marcu. 2006. Domain adaptation for statistical classifiers. Journal of Artifcial Intelligence Research 26 (2006), 101–126.

Huiming Duan, Xiaojing Bai, Baobao Chang, and Shiwen Yu. 2003. Chinese word segmentation at Peking University. In Proceedings of the second SIGHAN workshop on Chinese language processing. 152–155.

Zhen Guo, Yujie Zhang, Chen Su, and Jinan Xu. 2012. Exploration of N-gram Features for the Domain Adaptation of Chinese Word Segmentation. In Proceedings of Natural Language Processing and Chinese Computing Natural Language Processing and Chinese Computing. 121–131.

Wenbin Jiang, Liang Huang, and Qun Liu. 2009. Automatic Adaptation of Annotation Standards: Chinese Word Segmentation and POS Tagging - A Case Study. In Proceedings of ACL'09. 522–530.

Taku Kudo. 2013. CRF++: Yet Another CRF toolkit.

John Lafferty, Andrew McCallum, and Fernando Pereira. 2001. Conditional random fields: Probabilistic

models for segmenting and labeling sequence data. In Proceedings of ICML'01. 282–289.

Dekang Lin. 1998. Automatic Retrieval and Clustering of Similar Words. In Proceedings of ACL'98.

Yang Liu and Yue Zhang. 2012. Unsupervised Domain Adaptation for Joint Segmentation and POS Tagging. In Proceedings of COLING'12. 745–754.

Yijia Liu, Yue Zhang, Wanxiang Che, Ting Liu, and Fan Wu. 2014. Domain Adaptation for CRF-based Chinese Word Segmentation using Free Annotations. In Proceedings of EMNLP'14. 864–874.

Jin Kiat Low, Hwee Tou Ng, and Wenyuan Guo. 2005. A Maximum Entropy Approach to Chinese Word Segmentation. In Proceedings of the 4th SIGHAN Workshop on Chinese Language Processing. 970–979.

Hwee Tou Ng and Jin Kiat Low. 2004. Chinese Part-of-Speech Tagging: One-at-a-Time or All-at-Once? Word-Based or Character-Based? In Proceedings of EMNLP'04. 277–284.

Jerome Packard. 2000. The Morphology of Chinese: a cognitive and linguistic approach. Cambridge University Press.

Wenzhe Pei, Tao Ge, and Baobao Chang. 2014. Max-Margin Tensor Neural Network for Chinese Word Segmentation. In Proceedings of ACL'14. 293–303.

Fuchun Peng, Fangfang Feng, and Andrew McCallum. 2004. Chinese Segmentation and New Word Detection using Conditional Random Fields. In Proceedings of COLING'04.

Richard Sproat, Chilin Shih, William Gale, and Nancy Chang. 1996. A Stochastic Finite-State Word Segmentation Algorithm for Chinese. Computational Linguistics 22, 3 (1996), 377–404.

Weiwei Sun. 2010. Word-based and character-based word segmentation models: Comparison and combination. In Proceedings of ACL'10. 1211–1219.

Weiwei Sun. 2011. A Stacked Sub-Word Model for Joint Chinese Word Segmentation and Part-of-Speech Tagging. In Proceedings of ACL'11. 1385–1394.

Weiwei Sun and Jia Xu. 2011. Enhancing Chinese Word Segmentation Using Unlabeled Data. In Proceedings of EMNLP'11. 970–979.

Benjamin K. T'sou, Hing-Lung Lin, Godfrey Liu, Terence Chan, Jerome Hu, Ching hai Chew, and John K.P. Tse. 1997. A Synchronous Chinese Language Corpus from Different Speech Communities: Construction and Application. International Journal of Computational Linguistics and Chinese Language Processing 2, 1 (1997), 91–104.

Aobo Wang and Min-Yen Kan. 2013. Mining Informal Language from Chinese Microtext: Joint Word Recognition and Segmentation. In Proceedings of ACL'13. 731–741.

Fei Xia. 2000. The segmentation guidelines for the Penn Chinese Treebank (3.0).

Nianwen Xue. 2000. Defining and identifying words in Chinese. Ph.D. Dissertation. University of Delaware.

Nianwen Xue. 2003. Chinese Word Segmentation as Character Tagging. International Journal of Computational Linguistics and Chinese Language Processing 8, 1 (2003), 29–48.

Nianwen Xue, Fei Xia, Fu-Dong Chiou, and Martha Palmer. 2005. The Penn Chinese TreeBank: Phrase Structure Annotation of a Large Corpus. Natural Language Engineering 11, 2 (2005), 207–238.

Xiaodong Zeng, Derek F. Wong, Lidia S. Chao, and Isabel Trancoso. 2013a. Co-regularizing character-based and word-based models for semi-supervised Chinese word segmentation. In Proceedings of ACL'13. 171–176.

Xiaodong Zeng, Derek F. Wong, Lidia S. Chao, and Isabel Trancoso. 2013b. Graph-based Semi-Supervised Model for Joint Chinese Word Segmentation and Part-of-Speech Tagging. In Proceedings of ACL'13. 770–779.

Longkai Zhang, Li Li, Zhengyan He, Houfeng Wang, and Ni Sun. 2013a. Improving Chinese Word Segmentation on Micro-blog Using Rich Punctuations. In Proceedings of ACL'13. 177–182.

Longkai Zhang, Houfeng Wang, Xu Sun, and Mairgup Mansur. 2013b. Exploring Representations from Unlabeled Data with Co-training for Chinese Word Segmentation. In Proceedings of EMNLP'13. 311–321.

Meishan Zhang, Yue Zhang, Wanxiang Che, and Ting Liu. 2014. Type-Supervised Domain Adaptation for Joint Segmentation and POS-Tagging. In Proceedings of EACL'14. 588–597.

Yue Zhang and Stephen Clark. 2007. Chinese Segmentation Using a Word-based Perceptron Algorithm. In Proceedings of ACL'07. 840–847.

Hai Zhao, Chang-Ning Huang, and Mu Li. 2006. An improved Chinese word segmentation system with conditional random field. In Proceedings of the 5th SIGHAN Workshop on Chinese Language Processing. 162–165.

Xiaoqing Zheng, Hanyang Chen, and Tianyu Xu. 2013. Deep Learning for Chinese Word Segmentation and POS Tagging. In Proceedings of EMNLP'13. 647–657.